%% file: colomine.tex
\documentclass[runningheads]{llncs}

\usepackage{url}
\usepackage{amssymb,amsmath}
\usepackage[pdftex]{graphicx}
\usepackage[english]{babel}

\begin{document}	
	\title{Epoch-based Application of Problem-Aware Operators in a Multiobjective Memetic Algorithm for Portfolio Optimization 
        \thanks{This work is supported by Spanish Ministry of Economy under project Bio4Res (PID2021-125184NB-I00 -- \texttt{http://bio4res.lcc.uma.es}) and by Universidad de Málaga, Campus de Excelencia Internacional Andalucía Tech}
        }
	\titlerunning{Epoch-based Operator Application in a Multiobjective MA}
	
	\author{Feijoo Colomine Dur\'an\inst{1}\orcidID{0000-0002-2034-9205} \and \linebreak Carlos Cotta\inst{2,3}\orcidID{0000-0001-8478-7549} \and \linebreak Antonio J. Fern\'{a}ndez-Leiva\inst{2,3}\orcidID{0000-0002-5330-5217}}
	\authorrunning{F. Colomine et al.}
	\institute{Universidad Nacional Experimental del T\'{a}chira (UNET),
		Laboratorio de Computaci\'{o}n de Alto Rendimiento (LCAR), San
		Crist\'{o}bal, Venezuela \email{fcolomin@unet.edu.ve} \and
		Dept. Lenguajes y Ciencias de la Computaci\'on, ETSI Inform\'atica,
		Universidad de M\'alaga, Campus de Teatinos, 29071 - M\'alaga, Spain\\
		\email{\{ccottap,afdez\}@lcc.uma.es} \and
		ITIS Software, Universidad de Málaga, Spain }

	\maketitle
	
	\input{abstract}

	\input{introduction}

	\input{materialsAndMethods}

	\input{experimentation}

	\input{conclusions}

\input{biblio.tex}
\end{document}

%% file: abstract.tex
\begin{abstract}
\sloppypar We consider the issue of intensification/diversification balance in the context of a memetic algorithm for the multiobjective optimization of investment portfolios with cardinality constraints. We approach this issue in this work by considering the selective application of knowledge-augmented operators (local search and a memory of elite solutions) based on the search epoch in which the algorithm finds itself, hence alternating between unbiased search (guided uniquely by the built-in search mechanics of the algorithm) and focused search (intensified by the use of the problem-aware operators). These operators exploit Sharpe index (a measure of the relationship between return and risk) as a source of problem knowledge. We have conducted a sensibility analysis to determine in which phases of the search the application of these operators leads to better results. Our findings indicate that the resulting algorithm is quite robust in terms of parameterization from the point of view of this problem-specific indicator. Furthermore, it is shown that not only can other non-memetic counterparts be outperformed, but that there is a range of parameters in which the MA is also competitive when not better in terms of standard multiobjective performance indicators.
\end{abstract}

%% file: introduction.tex
\section{Introduction}
\label{sec:intro}
One of the crucial aspects for the successful application of metaheuristic optimization algorithms endowed with problem-aware search operators is the balance between intensification (the use of this knowledge to focus the search in particular search directions/regions) and diversification (a more exploratory behavior aimed to find solutions not perfectly aligned with the preferences dictated by the knowledge being exploited). This is particularly true in the case of multiobjective optimization scenarios, in which the output is generally desired to broadly cover the set of efficient solutions. We approach this issue in this work in the context of portfolio selection, and more precisely in the optimization of portfolios on the basis of their performance (the returns of the investment) and their risk (the inherent variance of share values for those assets included in the portfolio). 
There is a whole series of theoretical elements and studies related to the relationship
risk-return \cite{Philippe1999}, considering that the potential return or loss of investments is not static, but always depends on the evolution of the market. In this sense, the Markowitz model \cite{Markowitz1952}  has become a fundamental theoretical reference for the selection of investment portfolios (see Sect. \ref{sec:setting}), although its mathematical complexity (let alone the presence of different types of constraints) has sometimes caused its application in practice not to be as extensive. Of course, this also makes this area ripe for the application of multiobjective evolutionary algorithms (MOEAs) -- see \cite{6191315}. 

We approach the issue of intensification/diversification balance in the context of this problem via a memetic approach that features problem-aware operators (local search and a memory of elite solutions -- see Sect. \ref{sec:memetic}). These operators exploit a problem-specific indicator, namely the Sharpe index \cite{Sharpe1966}, a measure of the relationship between return and risk in a given investment portfolio. Our main goal is to (i) establish the usefulness of these operators, and (ii) to study which regime of application results in better performance, either when considered from the problem perspective (that is, the values attained for the problem-specific indicator) or from the point of view of standard indicators of multiobjective performance. To this end, we consider a simple strategy in which intensification happens during a specific time-frame of the execution of the algorithm. We have conducted a sensitivity analysis of the parameters governing the application of this strategy (Sect. \ref{sec:sensitivity}) as well as a comparison with other non-memetic MOEAs (Sect. \ref{sec:comparison}). Our findings indicate that despite its simplicity this strategy is flexible enough to allow modulating the intensification of the search so as to not only outperform other algorithms on the problem-specific indicator, but also to obtain competitive if not better results in terms of quality of the Pareto front attained.

%% file: materialsAndMethods.tex
\section{Materials and Methods}
\label{sec:materials}
As mentioned in Sect. \ref{sec:intro}, we have considered the use of memetic algorithms for the optimization of investment portfolios with cardinality restrictions in the context of the Markowitz model. We shall now describe in greater detail how the target problem is defined in this context, and how our algorithmic approach is structured.
\input{settings}

\input{memetic}

\input{data}

%% file: settings.tex
\subsection{The Optimization Problem}
\label{sec:setting}

The model developed by Markowitz \cite{Markowitz1952} aims to obtain a portfolio of $n$ assets that achieves the highest possible performance with the lowest risk. For this, average returns $R_i$ for each asset and the different covariances $\sigma_{ij}$ between the returns of assets (for $i=j$, this is $\sigma^2_i$, namely the variance of returns of the $i$-th asset) considered are calculated. Having performance and risk as the measures of interest, this model lends itself naturally to a multiobjective formulation as follows:
\smallskip
\begin{equation}
\min \sigma^{2}(\mathbf{w})= \sum_{i=1}^{n}\sum_{j=1}^{n}w_i w_j
\sigma_{ij}
\end{equation}
\begin{equation}
\max E[R(\mathbf{w})]= \sum_{i=1}^{n}w_iE(R_i) \label{equ}
\end{equation}
subject to
\begin{equation}
\sum_{i=1}^{n}w_i=1 
\end{equation}
where $\mathbf{w}=\langle w_1,\dots,w_n\rangle$, with $w_i\geqslant 0$ being the proportion of the investor's budget allocated to the $i$-th financial asset ($\mathbf{w}$ thus comprises the decision variables subject to optimization), $\sigma^{2}(\mathbf{w})$ is the variance of the portfolio given weights $\mathbf{w}$, and $E[R(\mathbf{w})]$ is the expected return of the portfolio defined by $\mathbf{w}$.

Some other restrictions can be added to this model, such as (i) cardinality, i.e., a maximum of $k$ weights is different from zero, 
\begin{equation}
\left|\{w_i > 0\ |\ 1\leqslant i \leqslant n\}\right| \leqslant k
\end{equation}
or allocation limits, i.e., any asset can take up to a maximum percentage $\rho$ of the portfolio
\begin{equation}
\forall i\in\{1,\dots,n\}:\  w_i \leqslant \rho.
\end{equation}
\sloppypar We herein consider the case of cardinality constraints. The set of pairs $[\sigma^{2}(\mathbf{w}),E[R(\mathbf{w})]]$ or risk-return combinations of all efficient portfolios (that is, of those portfolios in which the return cannot be increased without increasing the risk and vice versa, the risk cannot be reduced without having the return reduced as well) is called the efficient (Pareto) front ${\cal P}$. Once known, the investor, according to his preferences and the level of risk he is willing to assume, will choose his optimal portfolio \cite{Jin2021} from ${\cal P}$. while there is an obvious subjective component in the investor preferences, the financial literature also provides an objective performance measure that can be used by experts in order to quantify and compare how well a particular portfolio compares to another one. This measure is the Sharpe index \cite{Sharpe1966}, and it is defined as follows:     

\begin{equation}
S(\mathbf{w}) = \frac{{E[R(\mathbf{w})] - R_0 }}{{\sigma(\mathbf{w})}}\label{eq:sharpe}
\end{equation}
where $R_0$ is the assumed risk-free return. Thus, $E[R(\mathbf{w})] - R_0$ is the excess return obtained by taking some risk, and by dividing this quantity by $\sigma(\mathbf{w})$ (which is the standard deviation of the portfolio return, namely the square root of $\sigma^{2}(\mathbf{w})$), we obtain a measure of the excess return per unit of risk. Thus, higher values of this index indicate the portfolio performs better in the presence of risk. 

Quite interestingly, the Sharpe index also has an interesting geometric interpretation. The equation described in (\ref{eq:sharpe}) is a straight line that is tangent to the curve determined by the optimal set of portfolios ${\cal P}$ in the plane (using risk and return as horizontal and vertical coordinates respectively)  as a result of the solution of the Markowitz model.
Furthermore, given two portfolios $\mathbf{w_1}$ and $\mathbf{w_2}$, if the latter dominates (in a Pareto sense) the former --i.e., $E[R(\mathbf{w_1})] \leqslant E[R(\mathbf{w_2})]$ and $\sigma^{2}(\mathbf{w_1}) \geqslant \sigma^{2}(\mathbf{w_2})$, with at least one of the inequalities being strict-- then it must have a higher Sharpe index value (because $\mathbf{w_2}$ would have a larger numerator and a smaller denominator in Equation (\ref{eq:sharpe})). This suggests that the Sharpe ratio can be used within a multi-objective optimizer to move towards the Pareto front.

%% file: memetic.tex
\subsection{A Memetic Approach}
\label{sec:memetic}

As anticipated in Section \ref{sec:intro}, one of the key issues in successfully applying metaheuristics to a given optimization task is to endow the former with problem-aware components \cite{Wolpert1997NFL}. This is precisely one of the central tenets of memetic algorithms (MA) \cite{Moscato2019}. From a very specific point of view, these techniques arise from the combination of population-based optimization algorithms (often responsible for providing exploration/diversification capabilities) and some form of local search and/or problem-aware operators (which is in turn responsible for exploitation/search intensification in promising regions of the search space). While there are many other possibilities in the framework of memetic algorithms (e.g., see \cite{Neri2012c}), this basic skeleton suffices to build highly effective optimization methods. More importantly, these methods can be considered as a complementary problem-solving strategy, aimed at benefiting from existing algorithmic ideas, combining them synergistically. 

Given the multiobjective nature of the problem tackled, the population-based search component must be adapted to such an optimization scenario. To this end,  we have considered the use of the IBEA method \cite{Zitzler2004} as the underlying evolutionary search engine, due to its good performance in this problem domain \cite{ColomineDuran2012}. Following the problem description in Sect. \ref{sec:setting}, we have picked $[\sigma(\mathbf{w}), E[R(\mathbf{w})]]$ as the two objective functions. Solutions (i.e., the collection of weights corresponding to each of the funds in the portfolio) are represented as real-valued vectors.  After being subject to the variation operators --simulated binary crossover \cite{deb01SBX} (with parameter $\eta=0$) and polynomial mutation \cite{deb14pm} (with parameter $\eta_m=20$) in this case-- the cardinality constraint is enforced by picking the largest $k$ non-zero weights, setting the remaining entries to zero, and normalizing values so as to have adding to 1.0. As for the intensifying components, our algorithm incorporates two such elements:
\begin{itemize}
	\item \emph{Local search} (LS): solutions are subject to local improvement via the application of a first-ascent hill-climbing method with probability $P_{LS}$. This procedure works by mutating a single non-zero weight in the portfolio, and accepting the new solution if it has a better Sharpe index than the original solution (and this Sharpe index is also better than the population average ${\bar S}$). This is done until a certain fixed computational budget allocated to this component is exhausted.
	\item \emph{Elite memory} (EM): throughout the execution of the algorithm a memory of elite solutions is kept. This is an ordered list of the best individuals (according to the Sharpe index) generated by the algorithm in any given moment. This list is initially empty and has some fixed maximum size $\Theta$. Every time a solution is selected, it is incorporated to the elite memory if there is still available space in it; if the elite memory is at full capacity, the Sharpe index of the solution is checked against that of the worst solution currently in the elite memory, and the latter is substituted if worse. This elite memory is used as an intensification procedure as follows: if a selected solution is below the average Sharpe value of the population ${\bar S}$, a correction procedure is activated with probability $P_{EM}$, whereby the solution is substituted by a random member of the elite memory.
\end{itemize}

\begin{figure*}[!t]
	\begin{center}
		\includegraphics[width=\textwidth]{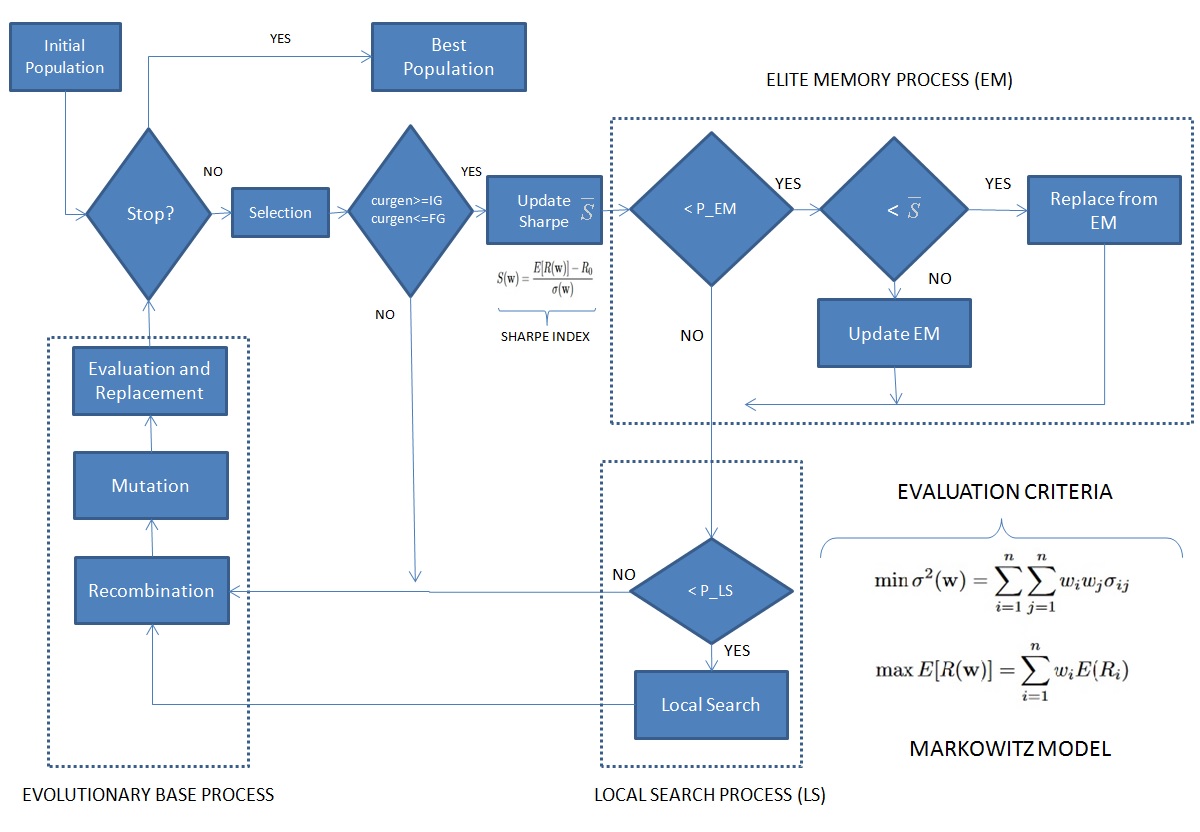}
	\end{center}
	\caption{Flowchart of the memetic approach considered}\label{fig:memetic}
\end{figure*}

While the adequate parameterization of these two components is an interesting topic in itself \cite{colomine21sensitivity}, there is another relevant factor to be considered in light of the underlying multiobjective scenario, namely the extent to which the application of these operators could narrow down the search too much towards specific direction of the optimal Pareto front. Indeed, obtaining good convergence and broad coverage of this optimal Pareto front remains a major challenge for many multiobjective metaheuristic optimization methods. To tackle this issue, we have considered the application of these two components within a fixed pre-determined time window during the execution of the algorithm. To be precise, we determine the initial generation $IG$ and the final generation $FG$ in which these operators are applied (see Figure \ref{fig:memetic}). The rationale behind this mechanism is to possibly allow the algorithm to expand its population broadly along the directions towards which the optimal Pareto front is located during an initial phase (which will be termed the \emph{deployment} stage), followed by an intensification stage whose duration is modulated by the particular values chosen for parameters $IG$ and $FG$), and in which the search prioritizes directions towards better values of the Sharpe index, and a final phase (which we shall term the \emph{securing} stage) in which the algorithm, whose population would be ideally close to at least some regions of the optimal Pareto front, advances in a Sharpe-unaware way in order to extend along the broadest collection of non-dominated solutions. Certainly, this is a procedure that admits further refinements and more elaborated strategies, but it serves as a first baseline to assess the potential use of this approach.

%% file: data.tex
\subsection{Data Used in the Analysis}
\label{sec:data}
The data considered for the experimental validation comprises the monthly closing prices in the Colombian Stock Market (\emph{Bolsa de Valores de Colombia} -- BVC) of twenty capital funds for years 2010 to 2016, obtained from the BVC website\footnote{The data is also available in our group's repository, at \url{https://osf.io/wg7mn/}.} \cite{Valores2016}. As a general rule, it is convenient to pick a time-frame for the data of at least five years, so as to --in an ideal scenario-- capture a full market cycle with the corresponding fluctuation in share prices. On the other hand, a very large time-frame (say, ten years or longer) is not advisable either, since the real value of the fund might not be appropriately represented by such old performances. In this sense, the dataset considered conveniently fits these considerations.

%% file: experimentation.tex
\section{Results}
\label{sec:experiments}
Having defined the algorithmic framework and data considered for the experimentation, we proceed to an experimental analysis of the MA in two phases: firstly, we shall conduct a sensitivity analysis to determine adequate time-frames for the intensification stage; then, we perform a comparative analysis of the best parameterizations of the MA with respect to other non-memetic MOEAs. This will be respectively done in Sect. \ref{sec:sensitivity} and Sect. \ref{sec:comparison} but prior to that, let us describe the computational setting for the experiments.

\input{configuration}

\input{results-sensitivity}

\input{results-comparison}

%% file: configuration.tex
\subsection{Experimental Setup}
\label{sec:configuration}

In addition to the MA, we have considered four non-memetic MOEAs, namely, NSGA-II \cite{NSGA2}, SPEA2 \cite{SPEA2}, SHV \cite{SHV} and HypE \cite{HypE}. The PISA library (a platform-independent interface and programming language for search algorithms) proposed by Bleuler et al. \cite{pisa} has been used for the base MOEAs. In all cases, the cardinality constraint is set to $k=18$, the crossover rate is $ P_X = 0.8$, the mutation rate is $ P_M = 0.005$, and the population size is $400$. With respect to the MA, elite memory has a size $\Theta =30$. Algorithms were run for a maximum number of 20,000 fitness evaluations --i.e., 50 generations-- (18,000 in the case of the MA to account for the overhead of local search), and 30 runs were performed for each of them.

To evaluate the results provided by each method, a double perspective has been considered, namely, the quality of the Pareto front obtained and the quality of the portfolio selected from this front. For the latter aspect, the Sharpe index is considered as mentioned in Section \ref{sec:setting}. More precisely, we pick a single portfolio from the Pareto front generated in each run using the Sharpe index as selection criterion. Regarding the evaluation of the Pareto fronts, two well-known performance indicators were used: the hypervolume indicator \cite{Zitzler1998} and the generational distance indicator (GD) \cite{Veldhuizen}. The former provides an indication of the region in the search space that is dominated by the front (thus, the bigger the better). This measure requires a reference point which in this case is defined by the maximum risk and minimum benefit observed in solutions from the best known Pareto front (the combined Pareto front obtained by all algorithms under comparison).
As for the second indicator, it estimates the extent to which a certain front is close to another (the true Pareto optimal front if known, or a reference front otherwise). Again, we take the combined Pareto front (that is, the Pareto front obtained by combining the fronts provided by all algorithms under comparison) as the reference set. Being a measure of distance to the reference set, the lower the value of GD, the better. In all cases (Sharpe index, hypervolume, GD), the final outcome of each batch of runs is thus a collection of performance values which will be subsequently subject to statistical analysis. All experimental results are available in our public data repository\footnote{\url{https://osf.io/qs8ae/}}.

%% file: results-sensitivity.tex
\subsection{Sensitivity Analysis} %
\label{sec:sensitivity}
In order to analyze the behavior of our MA in response to a different parameterization of the epoch-based intensification, we have conducted experiments in three different scenarios, namely, using either local search or elite memory in isolation ($P_{LS}=1, P_{EM}=0$ and $P_{LS}=0, P_{EM}=1$ respectively) , and using both components simultaneously ($P_{LS}=1, P_{EM}=1$). This provides a more detailed view of the effects that these parameters have on these components and any synergistic behavior that may show up. 

\begin{table}
\caption{\label{fig:gigf} Results of the MA for different values of $IG$ and $FG$. The best ($x^*$), median ($\tilde{x}$), quartile deviation (QD) and coefficient of quartile deviation (CQD) are shown for each performance indicator. The best configuration for each indicator is marked by $\star$; the remaining configurations are marked with symbols that indicate statistically significant difference at $\alpha = 0.1(\circ)$, $\alpha = 0.05(\bullet)$, and $\alpha = 0.01(${\tiny $\blacksquare$}) with this best value.}
\centering 
\resizebox{.905\textwidth}{!}{%
\begin{tabular}{ccrrrrcrrrrcrrrrc}
\hline
 &  & \multicolumn{15}{c}{$P_{LS} = 1, P_{EM} = 0$}\\
\cline{3-17}
 &  & \multicolumn{5}{c}{Sharpe index} & \multicolumn{5}{c}{hypervolume} & \multicolumn{5}{c}{GD}\\
\hline
$IG$  & $FG$  & $x^*$ & $\tilde{x}$ & QD & CQD & & $x^*$ & $\tilde{x}$ & QD & CQD & & $x^*$ & $\tilde{x}$ & QD& CQD & \\
\hline
0 & 10 & 0.408 & 0.406 & 0.001 & 0.139\% & {\tiny $\blacksquare$}\hspace{2mm} & 0.838 & 0.825 & 0.005 & 0.659\% & $\bullet$\hspace{2mm}  & 0.669 & 1.156 & 0.193 &  15.938\% & {\tiny $\blacksquare$}\hspace{2mm} \\
0 & 20 & 0.408 & 0.406 & 0.000 & 0.097\% & $\bullet$\hspace{2mm} & 0.839 & 0.830 & 0.005 & 0.560\% & \hspace{2mm}  & 0.641 & 0.896 & 0.084 &  9.281\% & $\star$\hspace{2mm} \\
0 & 30 & 0.408 & 0.407 & 0.000 & 0.118\% & \hspace{2mm} & 0.841 & 0.832 & 0.005 & 0.621\% & $\star$\hspace{2mm}  & 0.644 & 0.970 & 0.188 &  18.709\% & \hspace{2mm} \\
0 & 40 & 0.408 & 0.407 & 0.000 & 0.102\% & \hspace{2mm} & 0.839 & 0.831 & 0.003 & 0.362\% & \hspace{2mm}  & 0.668 & 0.907 & 0.160 &  16.794\% & \hspace{2mm} \\
0 & 50 & 0.408 & 0.407 & 0.000 & 0.071\% & $\star$\hspace{2mm} & 0.838 & 0.831 & 0.005 & 0.576\% & \hspace{2mm}  & 0.746 & 1.031 & 0.131 &  12.658\% & $\circ$\hspace{2mm} \\
10 & 20 & 0.407 & 0.405 & 0.000 & 0.116\% & {\tiny $\blacksquare$}\hspace{2mm} & 0.834 & 0.824 & 0.004 & 0.480\% & {\tiny $\blacksquare$}\hspace{2mm}  & 0.933 & 1.460 & 0.201 &  14.452\% & {\tiny $\blacksquare$}\hspace{2mm} \\
10 & 30 & 0.407 & 0.405 & 0.001 & 0.145\% & {\tiny $\blacksquare$}\hspace{2mm} & 0.837 & 0.825 & 0.002 & 0.286\% & {\tiny $\blacksquare$}\hspace{2mm}  & 0.878 & 1.315 & 0.140 &  10.421\% & {\tiny $\blacksquare$}\hspace{2mm} \\
10 & 40 & 0.407 & 0.406 & 0.001 & 0.134\% & {\tiny $\blacksquare$}\hspace{2mm} & 0.831 & 0.825 & 0.004 & 0.471\% & {\tiny $\blacksquare$}\hspace{2mm}  & 0.870 & 1.286 & 0.223 &  16.750\% & {\tiny $\blacksquare$}\hspace{2mm} \\
10 & 50 & 0.407 & 0.406 & 0.001 & 0.125\% & {\tiny $\blacksquare$}\hspace{2mm} & 0.837 & 0.825 & 0.004 & 0.446\% & $\bullet$\hspace{2mm}  & 0.829 & 1.335 & 0.212 &  15.758\% & {\tiny $\blacksquare$}\hspace{2mm} \\
20 & 30 & 0.405 & 0.404 & 0.001 & 0.168\% & {\tiny $\blacksquare$}\hspace{2mm} & 0.832 & 0.818 & 0.004 & 0.445\% & {\tiny $\blacksquare$}\hspace{2mm}  & 1.206 & 1.742 & 0.176 &  10.374\% & {\tiny $\blacksquare$}\hspace{2mm} \\
20 & 40 & 0.406 & 0.404 & 0.001 & 0.153\% & {\tiny $\blacksquare$}\hspace{2mm} & 0.831 & 0.817 & 0.003 & 0.331\% & {\tiny $\blacksquare$}\hspace{2mm}  & 1.003 & 1.682 & 0.206 &  12.107\% & {\tiny $\blacksquare$}\hspace{2mm} \\
20 & 50 & 0.406 & 0.404 & 0.001 & 0.220\% & {\tiny $\blacksquare$}\hspace{2mm} & 0.827 & 0.815 & 0.004 & 0.510\% & {\tiny $\blacksquare$}\hspace{2mm}  & 1.097 & 1.752 & 0.274 &  14.768\% & {\tiny $\blacksquare$}\hspace{2mm} \\
30 & 40 & 0.405 & 0.403 & 0.001 & 0.164\% & {\tiny $\blacksquare$}\hspace{2mm} & 0.822 & 0.809 & 0.005 & 0.623\% & {\tiny $\blacksquare$}\hspace{2mm}  & 1.489 & 2.093 & 0.299 &  14.522\% & {\tiny $\blacksquare$}\hspace{2mm} \\
30 & 50 & 0.405 & 0.403 & 0.001 & 0.149\% & {\tiny $\blacksquare$}\hspace{2mm} & 0.824 & 0.809 & 0.006 & 0.684\% & {\tiny $\blacksquare$}\hspace{2mm}  & 1.474 & 2.069 & 0.177 &  8.285\% & {\tiny $\blacksquare$}\hspace{2mm} \\
40 & 50 & 0.405 & 0.402 & 0.001 & 0.171\% & {\tiny $\blacksquare$}\hspace{2mm} & 0.822 & 0.804 & 0.007 & 0.847\% & {\tiny $\blacksquare$}\hspace{2mm}  & 1.378 & 2.209 & 0.314 &  14.724\% & {\tiny $\blacksquare$}\hspace{2mm} \\
\hline
 &  & \multicolumn{15}{c}{$P_{LS} = 0, P_{EM} = 1$}\\
\cline{3-17}
 &  & \multicolumn{5}{c}{Sharpe index} & \multicolumn{5}{c}{hypervolume} & \multicolumn{5}{c}{GD}\\
\hline
$IG$  & $FG$  & $x^*$ & $\tilde{x}$ & QD & CQD & & $x^*$ & $\tilde{x}$ & QD & CQD & & $x^*$ & $\tilde{x}$ & QD& CQD & \\
\hline
0 & 10 & 0.405 & 0.403 & 0.002 & 0.384\% & {\tiny $\blacksquare$}\hspace{2mm} & 0.824 & 0.806 & 0.011 & 1.348\% & {\tiny $\blacksquare$}\hspace{2mm}  & 1.178 & 1.987 & 0.373 &  18.412\% & {\tiny $\blacksquare$}\hspace{2mm} \\
0 & 20 & 0.409 & 0.408 & 0.000 & 0.057\% & $\star$\hspace{2mm} & 0.839 & 0.818 & 0.029 & 3.673\% & $\circ$\hspace{2mm}  & 0.594 & 1.005 & 0.493 &  37.962\% & $\star$\hspace{2mm} \\
0 & 30 & 0.409 & 0.408 & 0.000 & 0.089\% & $\bullet$\hspace{2mm} & 0.823 & 0.799 & 0.032 & 4.144\% & {\tiny $\blacksquare$}\hspace{2mm}  & 0.866 & 1.620 & 0.639 &  32.413\% & {\tiny $\blacksquare$}\hspace{2mm} \\
0 & 40 & 0.408 & 0.407 & 0.000 & 0.115\% & {\tiny $\blacksquare$}\hspace{2mm} & 0.802 & 0.762 & 0.014 & 1.786\% & {\tiny $\blacksquare$}\hspace{2mm}  & 1.986 & 2.662 & 0.278 &  10.570\% & {\tiny $\blacksquare$}\hspace{2mm} \\
0 & 50 & 0.408 & 0.407 & 0.000 & 0.115\% & {\tiny $\blacksquare$}\hspace{2mm} & 0.742 & 0.694 & 0.013 & 1.817\% & {\tiny $\blacksquare$}\hspace{2mm}  & 4.290 & 6.235 & 1.073 &  16.811\% & {\tiny $\blacksquare$}\hspace{2mm} \\
10 & 20 & 0.409 & 0.408 & 0.000 & 0.084\% & \hspace{2mm} & 0.841 & 0.817 & 0.024 & 2.950\% & \hspace{2mm}  & 0.536 & 1.273 & 0.350 &  30.365\% & \hspace{2mm} \\
10 & 30 & 0.408 & 0.407 & 0.000 & 0.080\% & {\tiny $\blacksquare$}\hspace{2mm} & 0.829 & 0.811 & 0.015 & 1.888\% & {\tiny $\blacksquare$}\hspace{2mm}  & 0.973 & 1.596 & 0.287 &  19.102\% & $\bullet$\hspace{2mm} \\
10 & 40 & 0.408 & 0.407 & 0.001 & 0.125\% & {\tiny $\blacksquare$}\hspace{2mm} & 0.799 & 0.755 & 0.009 & 1.232\% & {\tiny $\blacksquare$}\hspace{2mm}  & 1.965 & 2.774 & 0.243 &  9.004\% & {\tiny $\blacksquare$}\hspace{2mm} \\
10 & 50 & 0.408 & 0.407 & 0.001 & 0.147\% & {\tiny $\blacksquare$}\hspace{2mm} & 0.744 & 0.707 & 0.015 & 2.105\% & {\tiny $\blacksquare$}\hspace{2mm}  & 3.692 & 5.745 & 0.614 &  10.858\% & {\tiny $\blacksquare$}\hspace{2mm} \\
20 & 30 & 0.407 & 0.406 & 0.000 & 0.038\% & {\tiny $\blacksquare$}\hspace{2mm} & 0.834 & 0.826 & 0.003 & 0.409\% & $\star$\hspace{2mm}  & 0.769 & 1.309 & 0.168 &  12.737\% & \hspace{2mm} \\
20 & 40 & 0.407 & 0.406 & 0.000 & 0.107\% & {\tiny $\blacksquare$}\hspace{2mm} & 0.816 & 0.799 & 0.004 & 0.542\% & {\tiny $\blacksquare$}\hspace{2mm}  & 1.651 & 2.287 & 0.260 &  11.143\% & {\tiny $\blacksquare$}\hspace{2mm} \\
20 & 50 & 0.408 & 0.406 & 0.001 & 0.167\% & {\tiny $\blacksquare$}\hspace{2mm} & 0.752 & 0.730 & 0.010 & 1.349\% & {\tiny $\blacksquare$}\hspace{2mm}  & 3.665 & 5.217 & 0.815 &  15.920\% & {\tiny $\blacksquare$}\hspace{2mm} \\
30 & 40 & 0.407 & 0.406 & 0.001 & 0.130\% & {\tiny $\blacksquare$}\hspace{2mm} & 0.819 & 0.804 & 0.004 & 0.492\% & {\tiny $\blacksquare$}\hspace{2mm}  & 1.507 & 2.503 & 0.301 &  12.340\% & {\tiny $\blacksquare$}\hspace{2mm} \\
30 & 50 & 0.407 & 0.406 & 0.001 & 0.152\% & {\tiny $\blacksquare$}\hspace{2mm} & 0.765 & 0.746 & 0.012 & 1.658\% & {\tiny $\blacksquare$}\hspace{2mm}  & 3.608 & 5.114 & 0.610 &  11.248\% & {\tiny $\blacksquare$}\hspace{2mm} \\
40 & 50 & 0.407 & 0.405 & 0.001 & 0.184\% & {\tiny $\blacksquare$}\hspace{2mm} & 0.789 & 0.774 & 0.009 & 1.176\% & {\tiny $\blacksquare$}\hspace{2mm}  & 2.421 & 3.321 & 0.802 &  21.041\% & {\tiny $\blacksquare$}\hspace{2mm} \\
\hline
 &  & \multicolumn{15}{c}{$P_{LS} = 1, P_{EM} = 1$}\\
\cline{3-17}
 &  & \multicolumn{5}{c}{Sharpe index} & \multicolumn{5}{c}{hypervolume} & \multicolumn{5}{c}{GD}\\
\hline
$IG$  & $FG$  & $x^*$ & $\tilde{x}$ & QD & CQD & & $x^*$ & $\tilde{x}$ & QD & CQD & & $x^*$ & $\tilde{x}$ & QD& CQD & \\
\hline
0 & 10 & 0.409 & 0.409 & 0.000 & 0.037\% & {\tiny $\blacksquare$}\hspace{2mm} & 0.823 & 0.749 & 0.009 & 1.253\% & {\tiny $\blacksquare$}\hspace{2mm}  & 0.613 & 2.548 & 0.372 &  15.919\% & {\tiny $\blacksquare$}\hspace{2mm} \\
0 & 20 & 0.409 & 0.409 & 0.000 & 0.026\% & {\tiny $\blacksquare$}\hspace{2mm} & 0.767 & 0.746 & 0.003 & 0.372\% & {\tiny $\blacksquare$}\hspace{2mm}  & 2.063 & 2.679 & 0.104 &  3.922\% & {\tiny $\blacksquare$}\hspace{2mm} \\
0 & 30 & 0.409 & 0.409 & 0.000 & 0.028\% & {\tiny $\blacksquare$}\hspace{2mm} & 0.754 & 0.741 & 0.002 & 0.240\% & {\tiny $\blacksquare$}\hspace{2mm}  & 2.375 & 3.077 & 0.110 &  3.611\% & {\tiny $\blacksquare$}\hspace{2mm} \\
0 & 40 & 0.410 & 0.409 & 0.000 & 0.020\% & $\star$\hspace{2mm} & 0.721 & 0.710 & 0.003 & 0.431\% & {\tiny $\blacksquare$}\hspace{2mm}  & 4.677 & 5.609 & 0.279 &  5.047\% & {\tiny $\blacksquare$}\hspace{2mm} \\
0 & 50 & 0.409 & 0.409 & 0.000 & 0.022\% & {\tiny $\blacksquare$}\hspace{2mm} & 0.773 & 0.743 & 0.001 & 0.177\% & {\tiny $\blacksquare$}\hspace{2mm}  & 1.797 & 2.810 & 0.062 &  2.198\% & {\tiny $\blacksquare$}\hspace{2mm} \\
10 & 20 & 0.409 & 0.409 & 0.000 & 0.029\% & {\tiny $\blacksquare$}\hspace{2mm} & 0.820 & 0.749 & 0.006 & 0.807\% & {\tiny $\blacksquare$}\hspace{2mm}  & 0.673 & 2.530 & 0.162 &  6.395\% & {\tiny $\blacksquare$}\hspace{2mm} \\
10 & 30 & 0.409 & 0.409 & 0.000 & 0.023\% & {\tiny $\blacksquare$}\hspace{2mm} & 0.805 & 0.745 & 0.004 & 0.540\% & {\tiny $\blacksquare$}\hspace{2mm}  & 1.271 & 2.759 & 0.242 &  8.830\% & {\tiny $\blacksquare$}\hspace{2mm} \\
10 & 40 & 0.409 & 0.409 & 0.000 & 0.021\% & {\tiny $\blacksquare$}\hspace{2mm} & 0.720 & 0.708 & 0.005 & 0.645\% & {\tiny $\blacksquare$}\hspace{2mm}  & 4.583 & 5.538 & 0.491 &  8.651\% & {\tiny $\blacksquare$}\hspace{2mm} \\
10 & 50 & 0.409 & 0.409 & 0.000 & 0.022\% & {\tiny $\blacksquare$}\hspace{2mm} & 0.722 & 0.709 & 0.005 & 0.714\% & {\tiny $\blacksquare$}\hspace{2mm}  & 4.813 & 5.644 & 0.612 &  10.497\% & {\tiny $\blacksquare$}\hspace{2mm} \\
20 & 30 & 0.409 & 0.408 & 0.000 & 0.063\% & {\tiny $\blacksquare$}\hspace{2mm} & 0.840 & 0.830 & 0.004 & 0.490\% & $\star$\hspace{2mm}  & 0.781 & 1.245 & 0.302 &  23.308\% & $\star$\hspace{2mm} \\
20 & 40 & 0.409 & 0.408 & 0.000 & 0.047\% & {\tiny $\blacksquare$}\hspace{2mm} & 0.807 & 0.788 & 0.005 & 0.648\% & {\tiny $\blacksquare$}\hspace{2mm}  & 1.883 & 2.715 & 0.332 &  12.190\% & {\tiny $\blacksquare$}\hspace{2mm} \\
20 & 50 & 0.409 & 0.409 & 0.000 & 0.054\% & {\tiny $\blacksquare$}\hspace{2mm} & 0.776 & 0.745 & 0.011 & 1.417\% & {\tiny $\blacksquare$}\hspace{2mm}  & 3.076 & 5.284 & 1.165 &  20.992\% & {\tiny $\blacksquare$}\hspace{2mm} \\
30 & 40 & 0.408 & 0.407 & 0.000 & 0.081\% & {\tiny $\blacksquare$}\hspace{2mm} & 0.824 & 0.804 & 0.004 & 0.522\% & {\tiny $\blacksquare$}\hspace{2mm}  & 1.709 & 2.711 & 0.302 &  11.064\% & {\tiny $\blacksquare$}\hspace{2mm} \\
30 & 50 & 0.409 & 0.408 & 0.000 & 0.071\% & {\tiny $\blacksquare$}\hspace{2mm} & 0.775 & 0.757 & 0.005 & 0.638\% & {\tiny $\blacksquare$}\hspace{2mm}  & 4.307 & 6.009 & 1.093 &  17.833\% & {\tiny $\blacksquare$}\hspace{2mm} \\
40 & 50 & 0.408 & 0.407 & 0.001 & 0.135\% & {\tiny $\blacksquare$}\hspace{2mm} & 0.788 & 0.776 & 0.008 & 1.047\% & {\tiny $\blacksquare$}\hspace{2mm}  & 2.417 & 3.241 & 0.657 &  17.950\% & {\tiny $\blacksquare$}\hspace{2mm} \\
\hline
\end{tabular}%
}
\end{table}

The results are presented in the table \ref{fig:gigf}. Focusing firstly on the results for the configuration based on local search only, we can see that the best performing parameterization for all performance indicators are those that start with the intensification from the beginning ($IG = 0$). There is a difference in the duration of the intensification depending on the indicator though: for hypervolume and GD (that is, the pure multiobjective performance indicators), it seems that this intensification should not last until the latter stages of the run (there is a gentle degradation in the results of these indicators for larger values of $FG$, not enough in some cases to be statistically significant\footnote{Statistical significance is here determined with the Mann–Whitney U test \cite{wilcoxon} at the significance level $\alpha$ indicated in each case.}, although the trend seems to be present there); for the Sharpe index the best values are however obtained when local search is applied to the end of the algorithm. We can understand this behavior in terms of the pressure that local search exerts towards the region of the Pareto front containing high values of the Sharpe index. While deepening into this region can provide improvements in hypervolume and GD as well, it is clear that the latter also benefit from expanding the search toward Sharpe-suboptimal regions during the securing stage.

Let us now observe the results for the configuration based on the use of elite memory only. In this scenario all indicators seem to prefer a short intensification burst, albeit it must be noted that Sharpe index seems to benefit from an early application of the intensification, whereas hypervolume and GD also provide good results (the best ones for this configuration, or statistically undistinguishable form the best ones) when a short deployment phase is performed before the intensification stage. In this case, one has to take into account that the elite memory is a procedure whereby promising solutions generated in previous steps are retrieved and inserted back in the population. This has an obvious impact on the population diversity, which can be less detrimental during the initial stages in which diversity is still large. This is consistent with the degraded results obtained when this intensification only takes place at the end of the run: the population will be much closer to convergence and the benefits of reintroducing good known solutions are outweighed by the loss of diversity.

Finally moving to the joint use of local search and elite memory as intensification operators, the results are consistent with those of the previous configurations and seem to be not just more sharply defined, but also indicate both elements operate synergistically but without highly non-linear interactions. Thus, from the point of view of Sharpe index, the best configuration seems to be using a very low $IG$ (as in the case of both local search and elite memory alone), and a slightly lower value of $FG$ (larger than when elite memory is used on its own, but smaller than when using local search only). As to hypervolume and GD, these two indicators provide better results in the intermediate stages of the run (i.e., a deployment stage until $IG=20$, allowing the population to spread in all directions, followed by a short intensification until $FG=30$ that pushes forward the population towards the Sharpe-optimal region, and a final securing stage in which the population expands again from this bridgehead towards all directions of the Pareto front). From the point of view of absolute performance, it is clear that the combined used of both local search and elite memory is conducive to much improved results in terms of Sharpe index, at the cost of losing some coverage of other regions of the Pareto front. It is thus relevant to understand how this tradeoff is substantiated, and certainly whether the results can favorably compare to those of other MOEAs. This is done in next section.

%% file: results-comparison.tex
\subsection{Performance Comparison} %
\label{sec:comparison}
In order to gauge the performance of the MA with non-memetic MOEAs, we have conducted a comparative analysis. We have selected those parameterizations that provided the best results in terms of either of the performance indicators, as well as those for which no statistically significant difference was observed (not even at the $\alpha=0.1$ level). The notation MA$^{x,y}_{a,b}$ is used to denote the MA with $P_{LS}=x$, $P_{EM}=y$, $IG=a$, and $FG=b$. The results are presented in Table \ref{tab:comparison}.

\begin{table}[!t]
\caption{\label{tab:comparison} Results for the selected MA parameterization and non-memetic MOEAs. The best ($x^*$), median ($\tilde{x}$), quartile deviation (QD) and coefficient of quartile deviation (QCD) are shown for each performance indicator. The best configuration for each indicator is marked by $\star$; the remaining configurations are marked with symbols that indicate statistically significant difference at $\alpha = 0.1(\circ)$, $\alpha = 0.05(\bullet)$, and $\alpha = 0.01(${\tiny $\blacksquare$}) with this best value.}
\resizebox{\textwidth}{!}{
\begin{tabular}{crrrrcrrrrcrrrrc}
\cline{2-16}
 & \multicolumn{5}{c}{Sharpe index} & \multicolumn{5}{c}{hypervolume} & \multicolumn{5}{c}{GD}\\
\hline
algorithm  & $x^*$ & $\tilde{x}$ & QD & CQD & & $x^*$ & $\tilde{x}$ & QD & CQD & & $x^*$ & $\tilde{x}$ & QD& CQD & \\
\hline
MA$_{0,20}^{1,0}$ & 0.408 & 0.406 & 0.000 & 0.097\% & {\tiny $\blacksquare$}\hspace{2mm} & 0.839 & 0.830 & 0.005 & 0.560\% & \hspace{2mm}  & 0.641 & 0.896 & 0.084 &  9.281\% & $\star$\hspace{2mm} \\
MA$_{0,30}^{1,0}$ & 0.408 & 0.407 & 0.000 & 0.118\% & {\tiny $\blacksquare$}\hspace{2mm} & 0.841 & 0.832 & 0.005 & 0.621\% & $\star$\hspace{2mm}  & 0.644 & 0.970 & 0.188 &  18.709\% & \hspace{2mm} \\
MA$_{0,40}^{1,0}$ & 0.408 & 0.407 & 0.000 & 0.102\% & {\tiny $\blacksquare$}\hspace{2mm} & 0.839 & 0.831 & 0.003 & 0.362\% & \hspace{2mm}  & 0.668 & 0.907 & 0.160 &  16.794\% & \hspace{2mm} \\
MA$_{0,40}^{1,1}$ & 0.410 & 0.409 & 0.000 & 0.020\% & $\star$\hspace{2mm} & 0.721 & 0.710 & 0.003 & 0.431\% & {\tiny $\blacksquare$}\hspace{2mm}  & 4.677 & 5.609 & 0.279 &  5.047\% & {\tiny $\blacksquare$}\hspace{2mm} \\
MA$_{0,50}^{1,0}$ & 0.408 & 0.407 & 0.000 & 0.071\% & {\tiny $\blacksquare$}\hspace{2mm} & 0.838 & 0.831 & 0.005 & 0.576\% & \hspace{2mm}  & 0.746 & 1.031 & 0.131 &  12.658\% & $\circ$\hspace{2mm} \\
MA$_{20,30}^{1,1}$ & 0.409 & 0.408 & 0.000 & 0.063\% & {\tiny $\blacksquare$}\hspace{2mm} & 0.840 & 0.830 & 0.004 & 0.490\% & \hspace{2mm}  & 0.781 & 1.245 & 0.302 &  23.308\% & {\tiny $\blacksquare$}\hspace{2mm} \\
NSGA-II & 0.402 & 0.398 & 0.001 & 0.282\% & {\tiny $\blacksquare$}\hspace{2mm} & 0.808 & 0.782 & 0.006 & 0.772\% & {\tiny $\blacksquare$}\hspace{2mm}  & 2.256 & 3.408 & 0.543 &  15.377\% & {\tiny $\blacksquare$}\hspace{2mm} \\
SPEA2 & 0.401 & 0.398 & 0.002 & 0.407\% & {\tiny $\blacksquare$}\hspace{2mm} & 0.803 & 0.779 & 0.008 & 1.032\% & {\tiny $\blacksquare$}\hspace{2mm}  & 2.190 & 3.339 & 0.336 &  9.706\% & {\tiny $\blacksquare$}\hspace{2mm} \\
HypE & 0.406 & 0.405 & 0.000 & 0.115\% & {\tiny $\blacksquare$}\hspace{2mm} & 0.830 & 0.820 & 0.004 & 0.503\% & {\tiny $\blacksquare$}\hspace{2mm}  & 0.771 & 1.158 & 0.178 &  14.996\% & {\tiny $\blacksquare$}\hspace{2mm} \\
SHV & 0.392 & 0.380 & 0.004 & 1.115\% & {\tiny $\blacksquare$}\hspace{2mm} & 0.733 & 0.677 & 0.022 & 3.203\% & {\tiny $\blacksquare$}\hspace{2mm}  & 4.840 & 7.566 & 1.064 &  14.006\% & {\tiny $\blacksquare$}\hspace{2mm} \\
\hline
\end{tabular}}
\end{table}

\begin{figure}[!t]
\includegraphics[width=.32\textwidth]{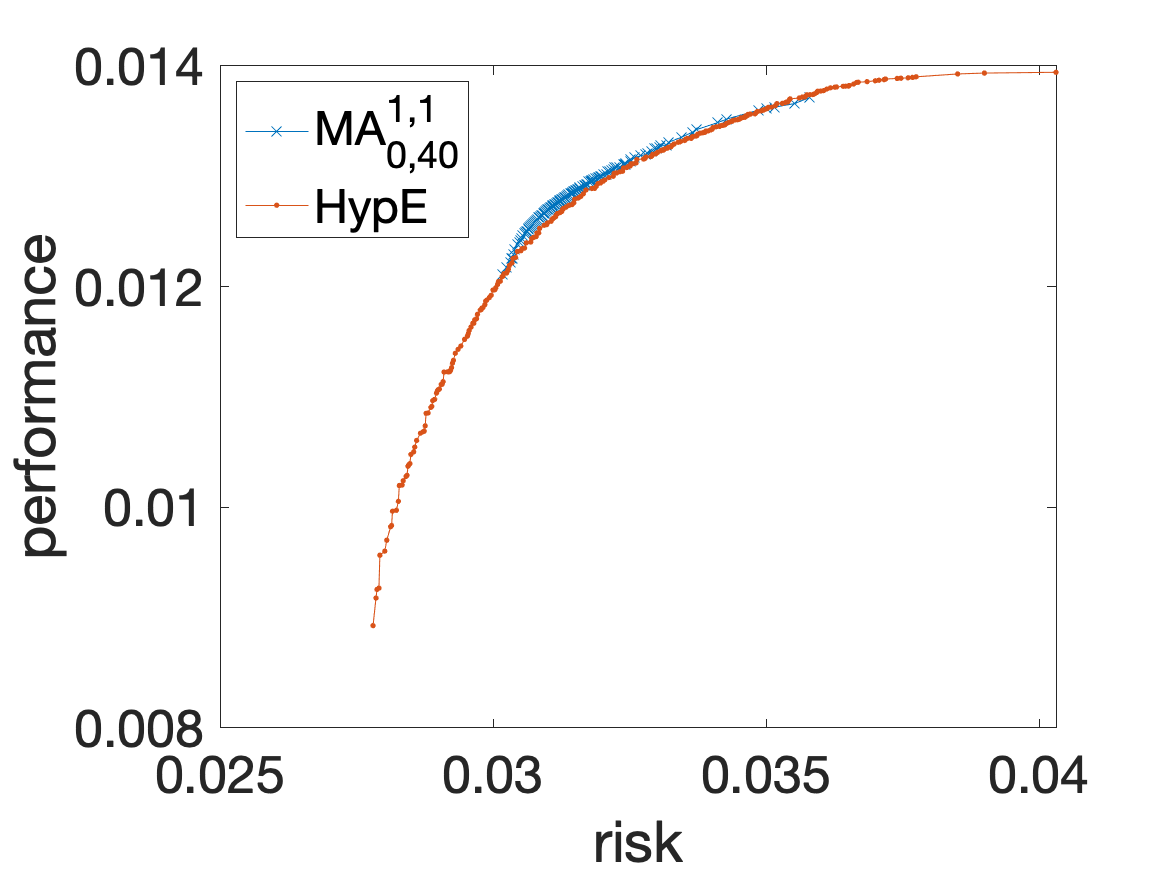}~\includegraphics[width=.32\textwidth]{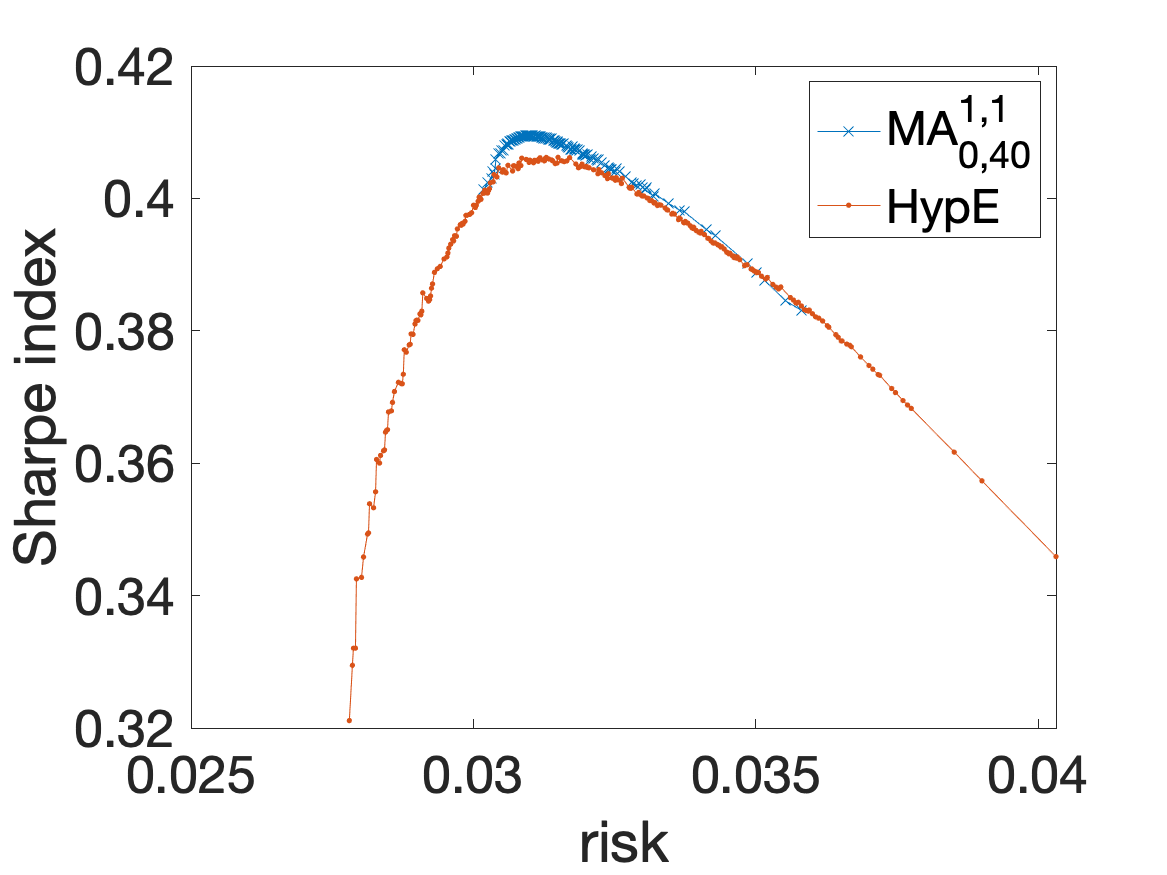}~\includegraphics[width=.32\textwidth]{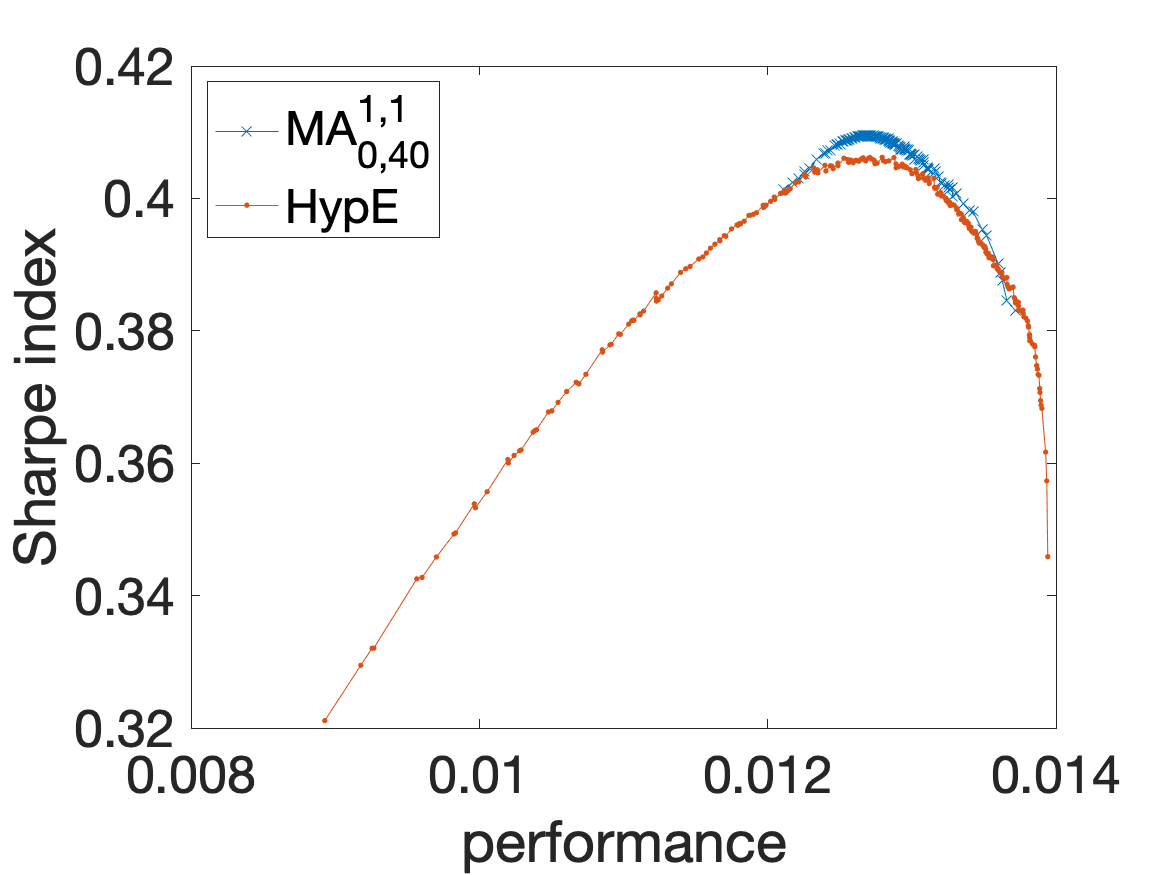}
\caption{\label{fig:fronts}Depiction of the combined Pareto fronts of MA$_{0,40}^{1,1}$ and HypE. The left plot shows the fronts achieved in the biobjective space. The middle and right plots show the Sharpe index for each solution in the front as a function of risk (middle) and performance (left).}
\end{figure}

As it can be seen, most MA parameterizations outperform the non-memetic MOEAs. This is particularly true for the Sharpe index indicator, for which MAs very significantly outperform the remaining algorithms. This is not surprising in light of the Sharpe-based intensification that takes place in the MAs, even if this intensification is done in different ways and with different parameterizations. It is nevertheless remarkable that in most cases, this superiority is also reflected in the standard multiobjective performance indicators. This indicates that progressing towards the Sharpe-optimal region of the Pareto front also provides very important advantages in terms of obtaining good fronts. This is illustrated in Fig. \ref{fig:fronts}, in which we depict the combined front (that is, the front attained by combining the Pareto fronts produced by the 30 runs of each algorithm) for MA$_{0,40}^{1,1}$ (the MA parameterization most oriented to Sharpe-index optimization) and HypE (the best performing MOEA in terms of hypervolume and GD). Notice firstly that in Fig. \ref{fig:fronts} (left), HypE seems to have a better coverage of extreme regions of the Pareto front (low risk/low performance and high risk/high performance), although MA$_{0,40}^{1,1}$ also extends towards the high-risk end of the front. The MA has a clear edge in the Sharpe-optimal region though, into which it achieves a deeper advance. This bulge is more clearly depicted in Fig. \ref{fig:fronts} (middle)-(right), showing the clear gain attained the MA in this region, which noticeably contributes to improving the multiobjective performance.

While having a better support of the Sharpe-optimal region of the Pareto front is obviously interesting from the point of view of the decision-maker, it is also interesting to note that there some parameterizations (e.g., MA$_{0, (30|40|50)}^{1,0}$) that despite providing slightly worse results in terms of the values of the Sharpe index attained, still show almost no statistically significant difference with the best performing algorithms in terms of hypervolume and GD. It is thus possible to tune the intensification stage so as to achieve a wider coverage of the Pareto front.

%% file: conclusions.tex
\section{Conclusions}
\label{sec:conclusions}

Finding the right balance between intensification and diversification is of paramount importance in general heuristic optimization, and turns out to be of particular interest in multiobjective endeavors in which the very existence of a target collection of efficient solutions provides substantial richness to the issue. In this sense, and focusing in this latter context, one has to keep in mind that the determination of the near-optimal Pareto front is not the end of the optimization process, but must be regarded in connection to the ulterior decision-making stage, whereby the expert will select an appropriate solution from the front. This is something that becomes very pertinent in portfolio optimization problems: not only do they constitute a usual scenario for the use of multi-objective evolutionary algorithms but they are also endowed with a natural indicator of solution efficiency (Sharpe index in this case). Among other things, this provides a legitimate measure to indicate preferred search directions in the multiobjective space. Indeed, exploiting this indicator can provide a reasonable search heuristic. The question thus arises as to how to integrate this heuristic into the search process, and how to best exploit it. We have approached the first question via an MA that features local search and a memory of elite solutions, always on the basis of the Sharpe index. As to the second issue, we have initially proposed a simple scheme in which intensification happens during a pre-specified time-window of the algorithm execution. This simple approach has nevertheless been able to show that it is possible to modulate the performance of the MA with respect to both Sharpe-based measures as well as to multiobjective performance indicators. In fact, not only is it generally possible to outperform non-memetic MOEAs on the basis of the former problem-aware measure, but the MA can be made competitive with these MOEAs in terms of pure multiobjective performance (hypervolume and GD in our case).  

The simplicity of the approach also paves the way for the consideration of more sophisticated strategies in the future. We believe that it would be of particular interest to determine an adaptive strategy, whereby the algorithm can adjust the use of these mechanisms during the run in response to population metrics and the state of the search. Needless to say, it would be convenient to extend these findings to other datasets or even to different domains.